%% file: acl2021.tex
\documentclass[11pt,a4paper]{article}
\usepackage[hyperref]{acl2021}
\usepackage{times}
\usepackage{latexsym}
\usepackage{pifont}
\usepackage{todonotes}
\usepackage{booktabs}
\usepackage{multirow}
\usepackage{amsmath}
\usepackage{amssymb}
\usepackage{colortbl}
\usepackage{xspace}
\usepackage{pifont}
\usepackage{graphicx}
\usepackage{textcomp}
\usepackage{xspace}
\usepackage{dsfont}
\usepackage{tabularx}
\usepackage{color, colortbl}
\usepackage{float}
\usepackage{textcomp}
\usepackage[inline]{enumitem}
\usepackage{wrapfig}

\definecolor{cthd}{rgb}{0.00392156862745098,0.45098039215686275,0.6980392156862745}
\definecolor{ct}{rgb}{0.00784313725490196,0.6196078431372549,0.45098039215686275}
\definecolor{lm}{rgb}{0.,0.,0.0}
\definecolor{pt}{rgb}{0.8,0.47058823529411764,0.7372549019607844}
\definecolor{ft}{rgb}{0.792156862745098,0.5686274509803921,0.3803921568627451}

\newcommand*\circled[1]{\tikz[baseline=(char.base)]{
            \node[shape=circle,draw,inner sep=2pt] (char) {#1};}}
\usepackage{microtype}

\newcommand{\modellm}{LM$_{\emptyset}$}
\newcommand{\modellmft}{S$_{\emptyset}$}
\newcommand{\modelsel}{S$_{\textrm{part}}$}
\newcommand{\modelfull}{S$_{\textrm{full}}$}

\newcommand{\compress}{\vspace{-2mm}}
\newcommand{\maptd}{\texttt{\small{FT}}}
\newcommand{\mapct}{\texttt{\small{CTX}}}
\newcommand{\maplm}{\texttt{\small{LM}}}
\newcommand{\mappt}{\texttt{\small{PT}}}
\newcommand{\mapcthd}{\texttt{\small{CTX-Hd}}}

\newcommand{\plm}{$P_{\text{\modellm}}$}
\newcommand{\plmft}{$P_{\emptyset}$ }

\newcommand{\msum}{M$_{\textrm{SUM}}$\xspace }
\newcommand{\mlm}{M$_{\textrm{LM}}$\xspace }
\newcommand{\xmark}{\text{\ding{55}}}
\newcommand{\cmark}{\text{\ding{51}}}

\aclfinalcopy 
\def\aclpaperid{3446} 

\title{Dissecting Generation Modes for Abstractive Summarization Models via Ablation and Attribution}

\author{Jiacheng Xu ~{\normalfont and}~ Greg Durrett \\
 Department of Computer Science \\
 The University of Texas at Austin\\
 {\tt \{jcxu,gdurrett\}@cs.utexas.edu}}

\date{}

\begin{document}
\maketitle
\begin{abstract}
Despite the prominence of neural abstractive summarization models, we know little about how they actually form summaries and how to understand where their decisions come from.
We propose a two-step method to interpret summarization model decisions. 
We first analyze the model's behavior by ablating the full model to categorize each decoder decision into one of several generation modes: roughly, is the model behaving like a language model, is it relying heavily on the input, or is it somewhere in between?
After isolating decisions that do depend on the input, we explore interpreting these decisions using several different attribution methods.
We compare these techniques based on their ability to select content and reconstruct the model's predicted token from perturbations of the input, thus revealing whether highlighted attributions are truly important for the generation of the next token.
While this machinery can be broadly useful even beyond summarization, we specifically demonstrate its capability to identify phrases the summarization model has memorized and determine where in the training pipeline this memorization happened, as well as study complex generation phenomena like sentence fusion on a per-instance basis.
\end{abstract}

\section{Introduction}
Transformer-based neural summarization models \cite{liu-lapata-2019-text,stiennon2020learning,xu-etal-2020-discourse,desai-etal-2020-compressive}, especially pre-trained abstractive models like BART \cite{lewis-2019-bart} and PEGASUS \cite{pegasus}, have made great strides in recent years. 
These models demonstrate exciting new capabilities in terms of abstraction, but little is known about how these models work. 
In particular, do token generation decisions leverage the source text, and if so, which parts? Or do these decisions arise based primarily on knowledge from the language model \cite{jiang-etal-2020-know,carlini2020extracting}, learned during pre-training or fine-tuning? Having tools to analyze these models is crucial to identifying and forestalling problems in generation, such as toxicity \cite{gehman2020realtoxicityprompts} or factual errors \cite{kryscinski-etal-2020-evaluating,goyal-durrett-2020-evaluating,goyal2020annotating}.


Although interpreting classification models for NLP has been widely studied from perspectives like feature attribution \cite{ribeiro2016should,intgrad} and influence functions \cite{koh2017understanding,han-etal-2020-explaining}, summarization specifically introduces some additional elements that make these techniques hard to apply directly. 
First, summarization models make sequential decisions from a very large state space. 
Second, encoder-decoder models have a special structure, featuring a complex interaction of decoder-side and encoder-side computation to select the next word.
Third, pre-trained LMs blur the distinction between relying on implicit prior knowledge or explicit instance-dependent input.


\input{fig_mot}

This paper aims to more fully interpret the step-wise prediction decisions of neural abstractive summarization models.\footnote{Code and visualization are available at \url{https://github.com/jiacheng-xu/sum-interpret}.}
First, we roughly bucket generation decisions into one of several \emph{modes} of generation. 
After confirming that the models we use are robust to seeing partial inputs, we can probe the model by predicting next words with various model \textbf{ablations}: a basic language model with no input (\modellm), a summarization model with no input (\modellmft), with part of the document as input (\modelsel), and with the full document as input (\modelfull).
These ablations tell us when the decision is context-independent (generated in an LM-like way), when it is heavily context-dependent (generated from the context), and more.
We \textit{map} these regions in  Figure~\ref{fig:map} and can use these maps to coarsely analyze model behavior. For example, 17.6\% of the decisions on XSum are in the lower-left corner (LM-like), which means they do not rely much on the input context.
Second, we focus on more fine-grained \textbf{attribution} of decisions that arise when the model \emph{does} rely heavily on the source document. We carefully examine interpretations based on several prior techniques, including occlusion \cite{zeiler2014visualizing}, attention, integrated gradients \cite{intgrad}, and input gradients \cite{inpgrad}. In order to evaluate and compare these methods, we propose a comprehensive evaluation based on presenting counterfactual, partial inputs to quantitatively assess these models' performance with different subsets of the input data.

Our two-stage analysis framework allows us to (1) understand how each individual decision depends on context and prior knowledge (Sec~\ref{sec:map}), (2) find suspicious cases of memorization and bias (Sec~\ref{sec:bias}), (3) locate the source evidence for context dependent generation (Sec~\ref{sec:content_attr}). The framework can be used to understand more complex decisions like sentence fusion (Sec~\ref{sec:fusion}). 
\section{Background \& Setup}
A seq2seq neural abstractive model first encodes an input document with $m$ sentences $(s_1, \cdots, s_m )$ and $n$ tokens $(w_{1}, w_{2}, \cdots, w_{n})$, then generates a sequence of tokens $(y_1, \cdots, y_T)$ as the summary. 
At each time step $t$ in the generation phase, the model encodes the input document and the decoded summary prefix and predicts the distribution over tokens as $p(y_t \mid w_1, w_2, \ldots, w_m, y_{<t} )$.

\subsection{Target Models \& Datasets}

We investigate the English-language CNN/DM \cite{hermann-2015-cnndm} and XSum \cite{narayan-2018-xsum} datasets, which are commonly used to fine tune pre-trained language models like BART, PEGASUS and T5. As shown in past work \cite{narayan-2018-xsum,chen-etal-2020-cdevalsumm,xu-etal-2020-understanding-neural}, XSum has significantly different properties from CNN/DM, so these datasets will show a range of model behaviors. We will primarily use the development sets for our analysis.

We focus on BART \cite{lewis-2019-bart}, a state-of-the-art pre-trained model for language modeling and text summarization. 
Specifically, we adopt `bart-large' as the language model \mlm, `bart-large-xsum' as the summarization model \msum for XSum, and `bart-large-cnn' for CNN/DM, made available by \citet{Wolf2019HuggingFacesTS}. 
BART features separate LM and summarization model sharing the same subword tokenization method.\footnote{Our analysis can generalize to other pre-trained models, but past work has shown BART and PEGASUS to be roughly similar in terms of behavior \cite{xu-etal-2020-understanding-neural}, so we do not focus on this here.}

Our approach focuses on teasing apart these different modes of decisions. We first run the full model to get the predicted summary $(y_1, \cdots, y_T )$. We then analyze the distribution placed by the full model \textbf{\modelfull{}} to figure out what contributes towards the generation of the next token. 



\subsection{Overview of Ablation and Attribution}
Figure~\ref{fig:main} shows our framework with an example of our analysis of four generation decisions. In the \textbf{ablation} stage, we compare the predictions of different model and input configurations. The goal of this stage is to coarsely determine the mode of generation.
Here, \emph{for} and \emph{Khan} are generated in an LM-like way: the model already has a strong prior that \emph{Sadiq} should be \emph{Sadiq Khan} and the source article has little impact on this decision. \emph{Cameron}, by contrast, does require the source in order to be generated. And \emph{mayoral} is a complex case, where the model is not strictly copying this word from anywhere in the source, but instead using a nebulous combination of information to generate it. In the \textbf{attribution} stage, we interpret such decisions which require more context using a more fine-grained approach. Given the predicted prefix (like \textit{David}), target prediction (like \textit{Cameron}), and the model, we use attribution techniques like integrated gradients \cite{intgrad} or LIME \cite{ribeiro2016should} to track the input which contributes to this prediction.

\subsection{Ablation Models and Assumptions}
\label{sec:partial_models}

The configurations we use are listed in Table~\ref{tab:models} and defined as follows:
\compress
\paragraph{\modellm} is a pre-trained language model only taking the decoded summary prefix as input. We use this model to estimate what a pure language model will predict given the prefix. We denote the prediction distribution as \plm$= P(y_t \mid y_{<t}; M_{\textrm{LM}} )$.

\begin{table}[t]
\centering
\footnotesize
\begin{tabular}{@{}r|cccc@{}}
\toprule
Config & \modellm & \modellmft & \modelsel & \modelfull \\ \midrule
Decoder prefix   &  \cmark  &  \cmark  & \cmark & \cmark  \\
Input document   &  \xmark  &  \xmark  & partial & full  \\
Model parameters & \mlm & \msum   & \msum &  \msum \\
\bottomrule
\end{tabular}
\caption{Model configurations with different amount of input document and back-end model. \mlm{} and \msum{} are the BART language model and summarization model respectively. \modellmft{} is the summarization model without any source document (encoder) input.}
\label{tab:models}
\end{table}

\compress
\paragraph{\modellmft} is the same BART summarization model as \modelfull{}, but without the input document as the input. That is, it uses the same parameters as the full model, but with no input document fed in. We use the prediction of this model to estimate how strong an effect the in-domain training data has, but still treating the model as a decoder-only language model. 
It is denoted as \plmft$= P(y_t \mid y_{<t}; M_{\textrm{SUM}} )$. Figure~\ref{fig:main} shows how this can effectively identify cases like \emph{Khan} that surprisingly do not rely on the input document.

\compress
\paragraph{\modelsel} is a further step closer to the full model: this is the BART summarization model conditioned on the decoder prefix and \emph{part of} the input document, denoted as 
$P_{\textrm{part}} = P(y_t \mid y_{<t}, \{s_i\}; M_{\textrm{SUM}})$ where $\{w_i\}$ is a subset of tokens of the input document.
The selected content could be a continuous span, or a sentence, or a concatenation of several spans or sentences. 



Although \msum is designed and trained to condition on input document, we find that the model also works well with no input, little input and incomplete sentences. As we will show later, there are many cases that this scheme successfully explains; we formalize our assumption as follows:
\compress
\paragraph{Assumption 1} \emph{If the model executed on partial input nearly reproduces the next word distribution of the full model, then we view that partial context as a \textbf{sufficient} (but perhaps not necessary) input to explain the model's behavior.}

Here we define \textit{partial input} as either just the decoded summary so far or the summary and partial context.
In practice, we see two things. First, when considering just the decoder context (i.e., behaving as an LM), the partial model may reproduce the full model's behavior (e.g., \emph{Khan} in Figure~\ref{fig:main}). We do not focus on explaining these cases in further detail. While conceivably the actual conditional model might internally be doing something different (a risk noted by \citet{rudin2019stop}), this proves the existence of a decoder-only proxy model that reproduces the full model's results, which is a criterion used in past work \cite{li-etal-2020-evaluating}. Second, when considering partial inputs, the model frequently requires one or two specific sentences to reproduce the full model's behavior, suggesting that the given contexts are both necessary \emph{and sufficient}.

Because these analyses involve using the model on data significantly different than that which it is trained on, we want another way to quantify the importance of a word, span, or sentence. This brings us to our second assumption:
\compress
\paragraph{Assumption 2} \emph{In order to say that a span of the input or decoder context is important to the model's prediction, it should be the case that this span is demonstrated to be important in counterfactual settings. That is, modified inputs to the model that include this span should yield closer predictions than those that don't.}

This criterion depends on the set of counterfactuals that we use. Rather than just word removal \cite{ribeiro2016should}, we will use a more comprehensive set of counterfactuals \cite{miller2019explanation,jacovi-goldberg-2020-towards} to quantify the importance of input tokens. We describe this more in Section~\ref{sec:content_attr}.

\subsection{Distance Metric}
\label{sec:distance}

Throughout this work, we rely on measuring the distance between distributions over tokens. Although KL divergence is a popular choice, we found it to be very unstable given the large vocabulary size, and two distributions that are completely different would have very large values of KL. We instead use the L$_1$ distance between the two distributions: $D(P, Q) = \sum_i |p_i - q_j|$.
This is similar to using the Earth Mover's Distance \cite{rubner1998metric} over these two discrete distributions, with an identity transportation flow since the distributions are defined over the same set of tokens.


\section{Ablation: Mapping Model Behavior}
\label{sec:map}


Based on Assumption 1, we can take a first step towards understanding these models based on the partial models described in Section~\ref{sec:partial_models}. Previous work \cite{see-2017-ptrgen,song2020controlling} has studied model behavior based on externally-visible properties of the model's generation, such as identifying novel words, differentiating copy and generation, and prediction confidence, which provides some insight about model's behavior \cite{xu-etal-2020-understanding-neural}. However, these focus more on shallow comparison of the input document, the generated summary, and the reference summary, and do not focus as strongly on the model.

We propose a new way of mapping the prediction space, with maps\footnote{While our axes are very different here, our mapping concept loosely follows that of \citet{swayamdipta-etal-2020-dataset}.} for XSum and CNN/DM shown in Figure~\ref{fig:map}. Each point in the map is a single subword token being generated by the decoder on the development set at inference time; that is, each point corresponds to a single invocation of the model. This analysis does not depend on the reference summary at all.

The $x$-axis of the map shows the distance between \modellm{} and \modelfull{}, using the metric defined in Section~\ref{sec:distance} which ranges from 0 to 2. The $y$-axis shows the distance between \modellmft{} and \modelfull{}. Other choices of partial models for the axes are possible (or more axes), but we believe these show two important factors. The $x$-axis captures \textbf{how much the generic pre-trained language model agrees with the full model's predictions}. The $y$-axis captures \textbf{how much the decoder-only summarization model agrees with the full model's predictions}. The histogram on the sides of the map show counts along with each vertical or horizontal slice.

\input{fig_map}


\paragraph{Modes of decisions} We break these maps into a few coarse regions based on the axis values. We list the coordinates of the bottom left corner and the upper right corner. These values were chosen by inspection and the precise boundaries have little effect on our analysis, as many of the decisions fall into the corners or along sides.
\compress
\paragraph{\maplm}{\footnotesize  ([0, 0], [0.5, 0.5])} contains the cases where \modellm{} and \modellmft{} both agree with \modelfull. These decisions are easily made using only decoder information, even without training or knowledge of the input document. These are cases that follow from the constraints of language models, including function words, common entities, or idioms.
\compress
\paragraph{\mapct}{\footnotesize ([0.5, 0.5], [2, 2])}  contains the cases where the input is needed to make the prediction: neither decoder-only model can model these decisions.
\compress
\paragraph{\maptd}{\footnotesize{([1.5, 0], [2, 0.5])}} captures cases where the fine-tuned decoder-only model is a close match but the pre-trained model is not. This happens more often on XSum and reflects memorization of training summaries, as we discuss later.
\compress
\paragraph{\mappt}{\footnotesize ([0, 1.5], [0.5, 2])} is the least intuitive case, where \modellm{} agrees with \modelfull{} but \modellmft{} does not; that is, fine-tuning a decoder-only model causes it to work \emph{less well}. This happens more often on CNN/DM and reflects memorization of data in the pre-training corpus.


\subsection{Coloring the Map with Context Probing}

While the map highlights some useful trends, there are many examples that do rely heavily on the context that we would like to further analyze. Some examples depend on the context in a sophisticated way, but other tokens like parts of named entities or noun phrases are simply copied from the source article in a simple way. Highlighting this contrast, we additionally subdivide the cases by how they depend on the context.


We conduct a sentence-level presence probing experiment to further characterize the generation decisions.
For a document with $m$ sentences, we run the \modelsel{} model conditioned on each of the sentences in isolation. We can obtain a sequence of scalars $P_{\textrm{sent}}=(P_{\textrm{part}}(s_i); i \in [1,m])$.
We define \mapcthd{} (``context-hard'') cases as ones where $\max(P_{\textrm{sent}})$ is low; that is, where no single sentence can yield the token, as in the case of sentence fusion. These also reflect cases of high entropy for \modelfull{}, where any perturbation to the input may cause a big distribution shift.
The first, second and third quartile of $\max (P_{\textrm{sent}})$ is $[0.69, 0.96, 1.0]$ and $[0.95, 1.0, 1.0]$ on XSum and on CNN/DM.




\subsection{Region Count \& POS Tags}
\input{tab_pos}

To roughly characterize the words generated in different regions of the map, in Table~\ref{tab:pos}, we show the percentage of examples falling to each region and the top 3 POS tags for each region on the XSum map. 
From the frequency of these categories, we can tell more than two-thirds of the decisions belong to the Context category. 17.6\% of cases are in LM, the second-largest category. 
In the LM region, ADP and DET account for nearly half of the data points, confirming that these are largely function words. Nouns are still prevalent, accounting for 13.5\% of the category. After observing the data, we found that these points represent commonsense knowledge or common nouns or entities, like ``Nations'' following ``United'' or ``Obama'' following ``Barack'' where the model generates these without relying on the input. 
Around 8\% of cases fall into gaps between these categories. Only 2.5\% and 2.1\% of the generations fall into the \mappt{} and \maptd{}, respectively. These are small but significant cases, as they clearly show the biases from the pre-training corpus and the fine-tuning corpus.  We now describe the effects we observe here.


\section{Bias from Training Data}
\label{sec:bias}
One benefit of mapping the predictions is to detect predictions that are suspiciously likely given one language model but not the other, specifically those in the \mappt{} and \maptd{} regions.
CNN/DM has more cases falling into \mappt{} than XSum so we focus on CNN/DN for \mappt{} and XSum for \maptd{}. 

\input{tab_bias_pt_mini}

\paragraph{\mappt{}: Bias from the Pretraining Corpus}
\label{subsec-pt}
The data points falling into the \mappt{} area are those where \modellm{} prediction is similar to \modelfull{} prediction but the \modellmft{} prediction is very different from \modelfull{}. 
We present a set of representative examples from the \mappt{} region of the CNN/DM map in Table~\ref{tab:ptlm-bias-mini}. 
For the first example, \emph{match} is assigned high probability by \modellm{} and \modelfull{}, but not by the no-input summarization models. The cases in this table exhibit a suspiciously high probability assigned to the correct answer in the base LM: its confidence about Kylie Jenner vs.~Kyle Min(ogue) is uncalibrated with what the ``true'' probabilities of these seem likely to be to our human eyes.

One explanation which we investigate is whether the validation and test sets of benchmark datasets like CNN/DM are contained in the pre-training corpus, which could teach the base LM these patterns.  Several web crawls have been used for different models, including C4 \cite{raffel2020exploring}, OpenWebText \cite{radford-2019-gpt2}, CC-News \cite{liu2019roberta}. Due to the availability of the corpus, we only check OpenWebText, which, as part of C4, is used for models like GPT-2, PEGASUS and T5.


According to \citet{hermann-2015-cnndm}, the validation and test sets of CNN/DM come from March and April 2015, respectively.
We extract the March to May 2015 dump of OpenWebText and find that 4.46\% (512 out of 11,490) test examples and 3.31\% (442 out of 13,368) validation examples are included in OpenWebText.\footnote{This is an approximation since we cannot precisely verify the pre-training datasets for each model, but it is more likely to be an underestimate than an overestimate. We only extract pre-training documents from \url{cnn.com} and \url{dailymail.co.uk} from a limited time range, so we may fail to detect snippets of reference summaries that show up in other time ranges of the scrape or in other news sources, whether through plagiarism or re-publishing.} 
Our matching criteria is more than three 7-gram word overlaps between the pre-training document and reference summaries from the dataset; upon inspection, over 90\% of the cases flagged by this criterion contained large chunks of the reference summary.

\input{tab_bias_td}

\textbf{Our conclusion is that the pre-trained language model has likely memorized certain articles and their summaries.} Other factors could be at play: other types of knowledge in the language model \cite{petroni-etal-2019-language,shin-etal-2020-autoprompt,oLMpics} such as key entity cooccurrences, could be contributing to these cases as well and simply be ``forgotten'' during fine-tuning. However, as an analysis tool, ablation suggested a hypothesis about data overlap which we were able to partially confirm, which supports its utility for understanding summarization models.




\paragraph{\maptd{}: Bias from Fine-tuning Data}

We now examine the data points falling in the bottom right corner of the map, where the fine-tuned LM matches the full model more closely than the pre-trained LM.

In Table~\ref{tab:bias-td}, we present some model-generated bigrams found in the \maptd{} region of XSum and compare the frequency of these patterns in the XSum and CNN/DM training data. Not every generation instance of these bigrams falls into the \maptd{} region, but many do. Table~\ref{tab:bias-td} shows the relative probabilities of these counts in XSum and CNN/DM, showing that these cases are all very common in XSum training summaries. The aggregate over all decisions in this region (the last line) shows this pattern as well. These can suggest larger patterns: the first three come from the common phrase \emph{in our series of letters from African journalists} (starts 0.5\% of summaries in XSum). Other stylistic markers, such as ways of writing currency, are memorized too.




\section{Attribution}
\label{sec:content_attr}

As shown in Table~\ref{tab:pos}, more than two thirds of generation steps actually do rely heavily on the context. 
Here, we focus specifically on identifying which aspects of the input are important for cases where the input \emph{does} influence the decision heavily using attribution methods. 



Each of the methods we explore scores each word $w_i$ in the input document with a score $\alpha_i$.
The score can be a normalized distribution, or a probability value ranging from 0 to 1.
For each method, we rank the tokens in descending order by score.
To confirm that the tokens highlighted are meaningfully used by the model when making its predictions, we propose an evaluation protocol based on a range of counterfactual modifications of the input document, taking care to make these compatible with the nature of subword tokenization.

\input{tab_increment}
\subsection{Evaluation by Adding and Removing}

Our evaluation focuses on the following question: given a budget of tokens or sentences, how well does the model reconstruct the target token $y_t$ when shown the important content selected by the attribution method? 
Our metric is the cross entropy loss of predicting the model-generated next token given different subsets of the input.\footnote{The full model is not a strict bound on this; restricting the model to only see salient content could actually increase the probability of what was generated. However, because we have limited ourselves to CTX examples and are aggregating across a large corpus, we do not observe this in our metrics.} 

Methods based on adding or removing single tokens have been used to evaluate before \cite{nguyen-2018-comparing}. However, for summarization, showing the model partial or ungrammatical inputs in the source may significantly alter the model's behavior. To address this, we use four methods to evaluate under a range of conditions, where in each case the model has a specific budget.
Our conditions are: 
\begin{enumerate*}
     \item \textsc{DispTok} selects $n$ tokens as the input.
    \item \textsc{RmTok} shows the document with $n$ tokens \textit{masked} instead of deleted.\footnote{Note that we do not directly remove the tokens because this approach typically makes the sentence ungrammatical. Token masks are a more natural type of input to models that are pre-trained with these sorts of masks anyway.}
    \item \textsc{DispSent} selects $n$ sentences as the input, based on cumulative attribution over the sentence.
    \item \textsc{RmSent} removes $n$ sentences from the document as the input. 
\end{enumerate*}

Table~\ref{tab:increment} shows examples of these methods applied to the examples from Figure~\ref{fig:main}. These highlight the impact of key tokens in certain generation cases, but not all.

We describe the details of how we feed or mask the tokens in \textsc{Tok} in Appendix.~\ref{app:detail-tok}.
The sentence-level methods are guaranteed to return grammatical input. Token-based evaluation is more precise which helps locating the exact feature token, but the trade-off is that the input is not fully natural.

\subsection{Methods}

We use two baseline methods: \textbf{Random}, which randomly selects tokens or sentences to display or remove, and \textbf{Lead}, which selects tokens or sentences according to document position, along with several attribution methods from prior work.
\textbf{Occlusion} \cite{zeiler2014visualizing} involves iteratively masking every single token or remove each sentence in the document and measuring how the prediction probability of the target token changes. 
Although \textbf{attention} has been questioned \cite{jain-wallace-2019-attention}, it still has some value as an explanation technique \cite{wiegreffe-pinter-2019-attention,serrano-smith-2019-attention}.
We pool the attention heads from the last layer of the Transformer inside our models, ignoring special tokens like SOS.

Finally, we use two gradient-based techniques \cite{bastings-filippova-2020-elephant}.  \textbf{Input Gradient} is a saliency based approach taking the gradient of the target token with respect to the input and multiplying by the input feature values. 
\textbf{Integrated Gradients} \citet{intgrad} computes gradients of the model input at a number of points interpolated between a reference ``baseline'' (typically an all-MASK input) and the actual input. This computes a path integral of the gradient. 



\paragraph{Attribution Aggregation for Sentence-level Evaluation}
We have described the six methods we use for token-level evaluation.
To evaluate these methods on the sentence level benchmark, we aggreagate the attributions in each sentence $attr(s_i) = \sum_{j=0}^d attr(w_j) / d$. 
Hence we can obtain a ranking of sentences by their aggregated attribution score.

\input{fig_bench}

\subsection{Results}

In Figure~\ref{fig:bench}, we show the token-level and sentence-level comparison of the attribution methods on the CTX examples in XSum.
IntGrad is the best technique overall, with InpGrad achieving similar performance. 
Interestingly, occlusion underperforms other techniques when more tokens are removed, despite our evaluation being based on occlusion; this indicates that single-token occlusion is not necessarily the strongest attribution method.
We also found that all of these give similar results, regardless of whether they present the model with a realistic input (sentence removal) or potentially ungrammatical or unrealistic input (isolated tokens added/removed).

Our evaluation protocol shows better performance from gradient-based techniques. The combination of four settings tests a range of counterfactual inputs to the model and increases our confidence in these conclusions.

\section{Case Study: Sentence Fusion}
\label{sec:fusion}

\input{tab_cases}

We now present a case study of the sort of analysis that can be undertaken using our two-stage interpretation method. We conduct an analysis driven by sentence fusion, a particular class of \mapcthd{} cases. Sentence fusion is an exciting capability of abstractive models that has been studied previously \cite{barzilay-mckeown-2005-sentence,thadani-mckeown-2013-supervised,lebanoff-etal-2019-analyzing,lebanoff-etal-2020-learning}.

We broadly identify cases of cross-sentence information fusion by first finding cases in \mapcthd{} where the $\max(P_{sent}) < 0.5$, but two sentences combined enable the model to predict the word. We search over all ${m \choose 2}$ combinations of sentences ($m$ is the total number of sentences) and run the \modelsel{} model on each pair of sentences. 
We identify 16.7\% and 6.0\% of cases in CNN/DM and XSum, respectively, where conditioning on a pair of sentences increases the probability of the model's generation by at least 0.5 over any sentence in isolation.

In Table~\ref{tab:fusion}, we show two examples of sentence fusion on XSum in this category, additionally analyzed using the \textsc{DispSent} attribution method. In the first example, typical in XSum, the model has to predict the event name \emph{UCI} without actually seeing it. The model's reasoning appears distributed over the document: it consults entity and event descriptions like \emph{world champion} and \emph{France}, perhaps to determine this is an international event.
In the second example, we see the model again connects several pieces of information. The generated text is factually incorrect: the horse is retiring, and not Dujardin. Nevertheless, this process tells us some things that are going wrong (the model disregards the horse in the generation process), and could potentially be useful for fine-grained factuality evaluation using recent techniques \cite{tian2019sticking,kryscinski-etal-2020-evaluating,goyal-durrett-2020-evaluating,maynez-etal-2020-faithfulness}.

The majority of the ``fusion'' cases we investigated actually reflect content selection at the beginning of the generation. Other cases we observe fall more cleanly into classic sentence fusion or draw on coreference resolution.



\section{Related Work}
Model interpretability for NLP has been intensively studied in the past few years \cite{ribeiro2016should,AMJ,jacovi-etal-2018-understanding,chen-etal-2020-generating,jacovi-goldberg-2020-towards,deyoung-etal-2020-eraser,pruthi-etal-2020-learning,ye2021evaluating}. However, many of these techniques are tailored to classification tasks like sentiment. For post-hoc interpretation of generation, most work has studied machine translation \cite{maanalysis,li-etal-2020-evaluating,voita2020analyzing}. \citet{li-etal-2020-evaluating} focus on evaluating explanations by finding surrogate models that are similar to the base MT model; this is similar to our evaluation approach in Section~\ref{sec:content_attr}, but involves an extra distillation step. Compared to \citet{voita2020analyzing}, we are more interested in highlighting how and why changes in the source article will change the summary (\emph{counterfactual explanations}).

To analyze summarization more broadly, \citet{xu-etal-2020-understanding-neural} provides a descriptive analysis about models via uncertainty.  
Previous work \cite{kedzie-etal-2018-content,zhong-etal-2019-searching,kryscinski-etal-2019-neural,zhong-etal-2019-searching} has conducted comprehensive examination of the limitations of summarization models. 
\citet{filippova-2020-controlled} ablates model input to control the degree of hallucination. \citet{miao2021prevent} improves the training of MT by comparing the prediction of LM and MT model.


Finally, this work has focused chiefly on abstractive summarization models. We believe interpreting extractive \cite{liu-lapata-2019-text} or compressive \cite{xu-durrett-2019-neural,xu-etal-2020-discourse,desai-etal-2020-compressive} models would be worthwhile to explore and could leverage similar attribution techniques, although ablation does not apply as discussed here.

\section{Recommendations \& Conclusion}

We recommend a few methodological takeaways that can generalize to other conditional generation problems as well.

First, \textbf{use ablation to analyze generation models.} While removing the source forms inputs not strictly on the data manifold, ablation was remarkably easy, robust, and informative in our analysis. Constructing our maps only requires querying three models with no retraining required.

Second, to understand an individual decision, \textbf{use feature attribution methods on the source \emph{only}.} Including the target context often muddies the interpretation since recent words are always relevant, but looking at attributions over the source and target together doesn't accurately convey the model's decision-making process.


Finally, to probe attributions more deeply, \textbf{consider adding or removing various sets of tokens.} The choice of counterfactuals to explain is an ill-posed problem, but we view the set used here as realistic for this setting \cite{ye2021evaluating}.

Taken together, our two-step framework allows us to identify generation modes and attribute generation decisions to the input document. Our techniques shed light on possible sources of bias and can be used to explore phenomena such as sentence fusion. We believe these pave the way for future studies of targeted phenomena, including fusion, robustness, and bias in text generation, through the lens of these interpretation techniques.

\section*{Acknowledgments}

Thanks to the members of the UT TAUR lab for helpful discussion, especially Tanya Goyal, Yasumasa Onoe, and Xi Ye for constructive suggestions.
This work was partially supported by a gift from Salesforce Research and a gift from Amazon. Thanks as well to the anonymous reviewers for their helpful comments.


\bibliographystyle{acl_natbib}
\bibliography{acl2021}

\clearpage
\appendix

\input{fig_map_cnndm}
\begin{table}[t]
\centering
\footnotesize
\begin{tabular}{r|ccc} \toprule
            & Complexity                                         & Time & Memory  \\ \midrule
Occlusion   & $\mathcal{O}(n^3)$                          & $\sim$ 33x  &  $\sim$ 2.1x     \\
S+Occlusion & $\mathcal{O}(s \times d^2 + d^3)$      & 1x   &   1x   \\ \bottomrule
\end{tabular}
\begin{tabular}{r|ccccc}
\toprule
\multicolumn{1}{c|}{\textsc{DispTok}} & 0                     & 1    & 2    & 4    & 8    \\ \midrule
Occlusion            &           \multirow{2}{*}{4.61}           & 4.28 & 3.97 & 3.36 & 2.84 \\
S+Occlusion      &  & 4.27 & 3.93 & 3.31 & \textbf{2.71} \\
\bottomrule
\end{tabular}
\caption{(Upper) The complexity, actual time and GPU memory comparison of Occlusion and S+Occlusion. We set the same environment for both experiments. (Bottom) Token-level selection evaluation on Occlusion and S+Occlusion. The reported number is the NLL loss of w.r.t. the token predicted by \modelfull.
}
\label{tab:occ}
\end{table}

\input{fig_color_example}

\section{Validity of Decoder-Only Model in \modellmft{} Setting}
We use an off-the-shelf BART summarization model as the decoder-only model for the ablation study. 
To guarantee the validity of the usage of the off-the-shelf model for ablation study, we also \textit{fine-tuned} a BART language model where encoding input is empty and the decoding target is the reference summary.
We compare the model output with the \modellmft{} output in the paper. For 55\% of cases the top-1 predictions of these two models agree with each other. This is pretty high, and suggests that the \modellmft{} is at least doing reasonably. Note that fine-tuning will probably give rise to different behavior on the ~70\% of CTX cases, since the \modellmft{} will hallucinate differently than the newly fine-tuned model (which further suggests why our analysis should focus on \modellmft{}).

\section{Examples of \mappt{}}
We present more examples of bias from the pre-trained language model on CNN/DM in Table~\ref{tab:ptlm-bias}. In Table~\ref{tab:ptlm-bias-mini} we have shown the cases where the memorized phrases are proper nouns or nouns. Here we provide examples of other types like function words. 
The memorization of function words like \emph{with} or \emph{and} can be challenging to spot using other means due to their ubiquity. 
\input{tab_bias_pt}

\input{tab_tok}

\input{tab_sent}

\section{Implementation Detail for \textsc{Tok}}
\label{app:detail-tok}
We rank the attribution score of all subword tokens rather than words. However, to provide necessary context for \textsc{DispTok} and to avoid information leakage in \textsc{RmTok}, we extend the selection by a context window to collect neighboring word pieces. We illustrate the way of fulfilling budget with an example.
\begin{table}[H]
\centering
\footnotesize
\begin{tabular}{ccccccc}
\textbf{Bur} & \#berry &  bets &  on &  \textbf{new} &  branding  \\
\circled{1} & \circled{2} &  & \circled{4} & \circled{3} & \textcolor{gray}{\circled{5}}  
\end{tabular}
\label{tab:budget}
\end{table}
In this example ``Bur'' receives the highest score and ``new'' the second. We use a context windows of size 1 and a budget of $n=4$ tokens. 
In \textsc{DispTok}, the input will be ``\textlangle{}sos\textrangle{}Burberry, on new\textlangle{}eos\textrangle{}''; 
In \textsc{RmTok}, the input will be \textlangle{}sos\textrangle{}\#\# bets\#\# branding\textlangle{}eos\textrangle{} where \# stands for the MASK token. If $n=5$, \emph{branding} will be added or masked.

\section{Efficient Two-Stage Selection Model}

For long documents in summarization, attribution methods can be computationally expensive. Occlusion requires running inference once for each token in the input document. Gradient-based methods store the gradients and so require a lot of GPU memory when the document is long. These techniques spend time and memory checking words that have little impact on the generation.

In order to improve the efficiency of these methods, we propose an efficient alternative where we first run sentence level presence probing on the full document, and then run attribution methods locally on the top-$k$ sentences. 
We call the proposed model \emph{S+[method]} where \emph{method} can be arbitrary attribution methods including occlusion, attention, InpGrad and IntGrad.

We define our notation as follows:
$s$, $n$ and $d$ are the number of sentences, the number of tokens in the document, and the number of tokens in each sentence, respectively. 
For the occlusion method, we can run inference $s$ times to pre-select important sentences, each of which costs $\mathcal{O}(d^2)$ times due to self-attention. 
The attribution is then applied only to only one or few sentences so the complexity is now $\mathcal{O}(k \times d^2 \times d)$ where $k$ is the number of top sentences used for attribution. In our experiments, we set $k=2$ and $n\leq500$. 
Compared to the complexity of the regular model $\mathcal{O}(n^3)$, the complexity of the two-stage model is only $\mathcal{O}(s \times d^2 + k \times d^2 \times d)$. 
\input{tab_complex}

In Table~\ref{tab:occ} we compare the complexity and actual run time and memory usage. We batch the occlusion operation and the batch size is set to 100. We can see a huge reduction in running time and a significant drop in memory usage. 

\paragraph{Takeaway} 
A two-stage selection model is much more efficient, yielding a 97\% running time reduction on the occlusion method. The downside of this method is that it only produces single-sentence attributions, and so isn't appropriate in cases involving sentence fusion.

Following \citep{vaswani-2017-transformer}, we compare the complexity for all methods in Table~\ref{tab:complex}.
$n$ is the number of tokens in the document. $d$ is the number of tokens in each sentence. $s$ is the number of sentences in the document.  $r$ is the number of steps in the integral approximation of Integrated Gradient. $bp$ indicates the time consumption of one back-propagation for gradient based methods. We list the complexity of the original methods in the middle column and the sentence based pre-selection variant in the right column. The base cost for sentence pre-selection model is to run the sentence selection model $s$ times, so it's $\mathcal{O}(s \times d^2)$. 
The $n^2$ and $d^2$ originate from the quadratic operation of self-attentions in Transformer models. We ignore the number of layers in the neural network or other model related hyper-parameters since all of the methods here share the same model. 

\section{Four Way Evaluation}
Due to the space limit, we only show the plot of the four way evaluation in Figure~\ref{fig:bench}.
To enable future comparisons on the proposed evaluation protocol, we also include the detailed results in Table~\ref{tab:main-tok} and Table~\ref{tab:main-sent} for \textsc{Tok} and \textsc{Sent} evaluation. The $\Delta$ measures how the average performance increase or drop deviates from the original baseline. We abstract the evaluation methods as a function $eval$. The input is the text and the budget $n$ and output is the predicted loss.
\begin{align*}
    \Delta = \textrm{Avg}(eval(i)) - eval(0) 
\end{align*}

For \textsc{Tok} series evaluation, $i \in \{1, 2, 4, 8, 16\}$.
For \textsc{Sent} series evaluation, $i \in \{1, 2, 3, 4\}$ because a sentence carries much more information than a token. 
IntGrad performs the best across all of the evaluation methods. 
\end{document}

%% file: fig_mot.tex
\begin{figure*}[t]
\centering
\small
\includegraphics[width=1\textwidth]{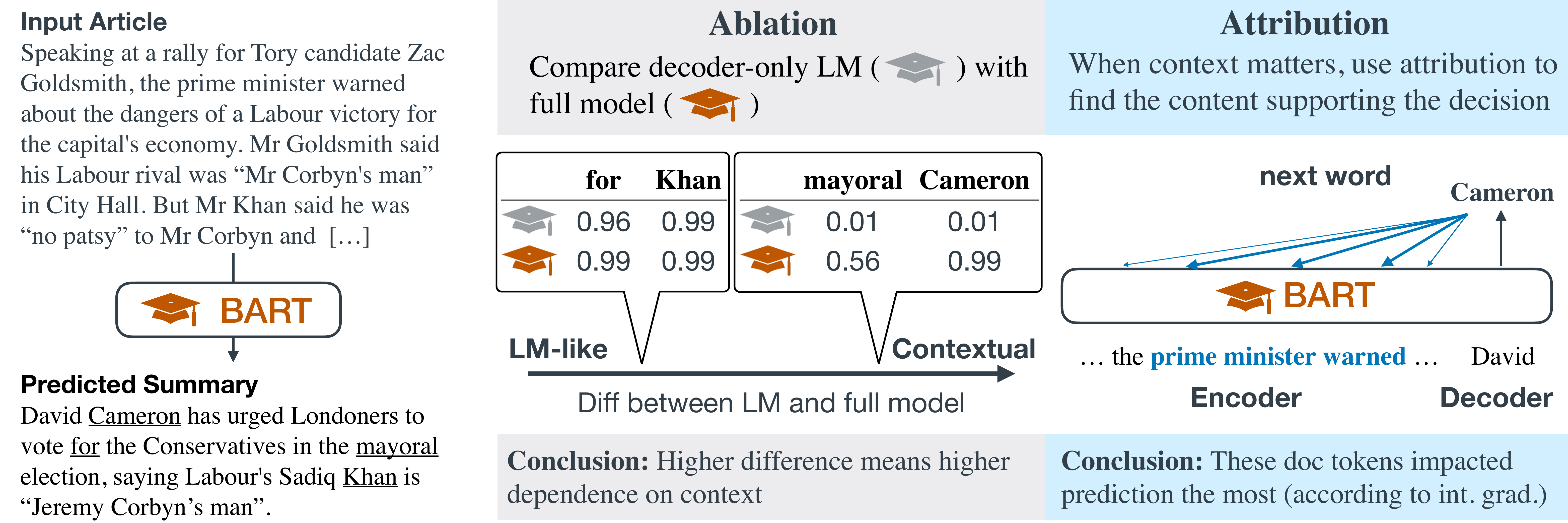}

\caption{Our two-stage ablation-attribution framework. First, we compare a decoder-only language model (not fine-tuned on summarization task, and not conditioned on the input article) and a full summarization model. They are colored in gray and orange respectively. the The higher the difference, the more heavily model depends on the input context. For those context-dependent decisions, we conduct content attribution to find the relevant supporting content with methods like Integrated Gradient or Occlusion. 
}
\label{fig:main}
\end{figure*}


%% file: fig_map.tex
\begin{figure}[t]
\centering
\small
\includegraphics[width=0.38\textwidth]{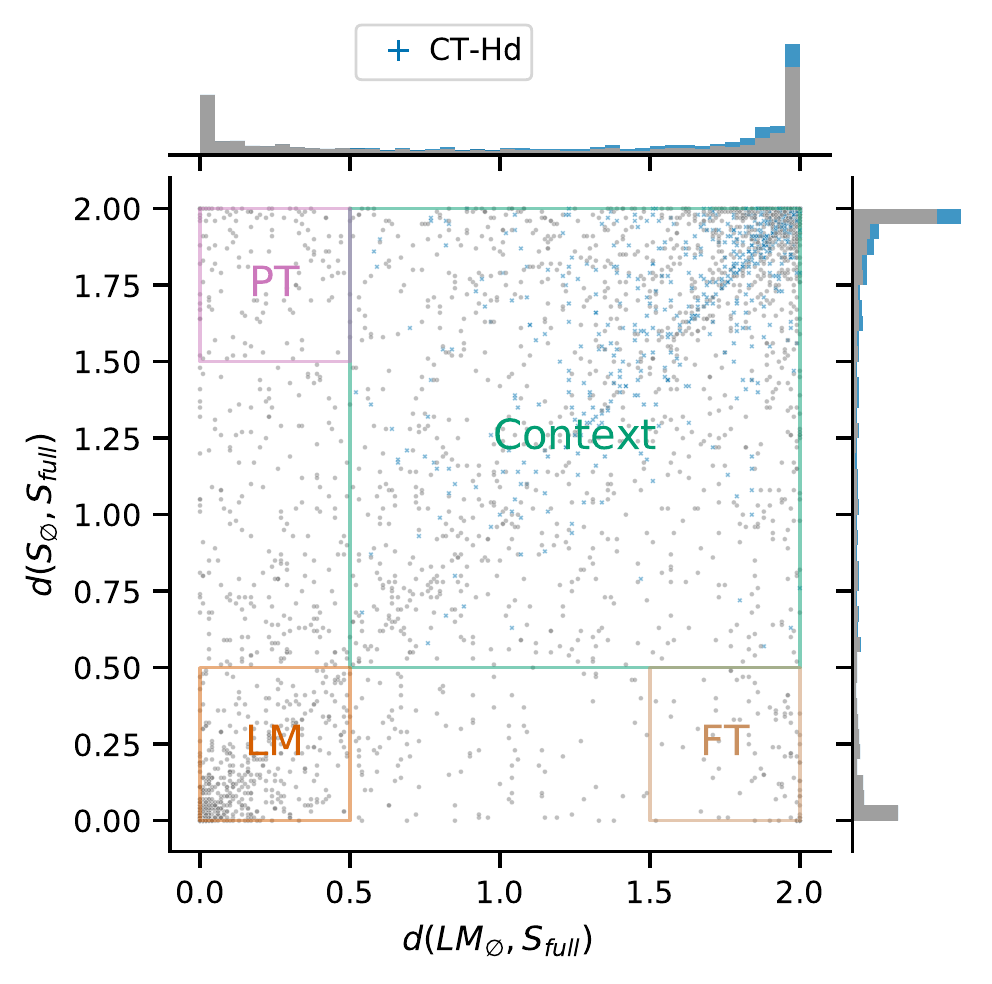}
\includegraphics[width=0.38\textwidth]{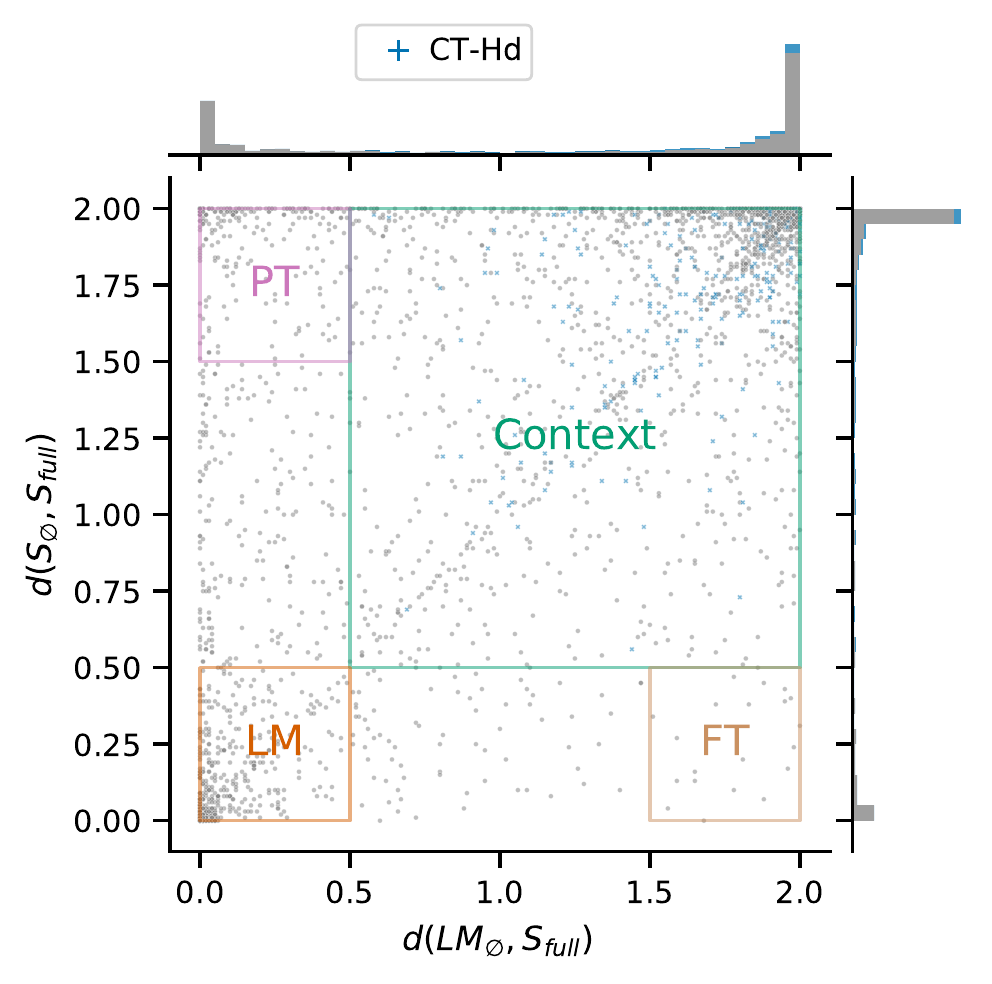}
\caption{Map of model behavior on XSum (top) and CNN/DM (bottom). The $x$-axis and $y$-axis show the distance between \modellm{} and \modelfull{}, and distance between \modellmft{} and \modelfull{}. The regions characterize different generation modes, defined in Section~\ref{sec:map}.}
\label{fig:map}
\end{figure}



%% file: tab_pos.tex
\begin{table}[t]
\centering
\small
\begin{tabular}{
>{\columncolor[HTML]{FFFFFF}}c 
>{\columncolor[HTML]{FFFFFF}}c 
>{\columncolor[HTML]{FFFFFF}}c 
>{\columncolor[HTML]{FFFFFF}}c 
>{\columncolor[HTML]{FFFFFF}}c }
\toprule
Cat                                          & Freq(\%)                                       & \multicolumn{3}{c}{\cellcolor[HTML]{FFFFFF}Top 3 POS Tags w/ Freq(\%)} \\\midrule
\cellcolor[HTML]{FFFFFF}                     & \cellcolor[HTML]{FFFFFF}                        & ADP                    & DET                    & NOUN                  \\
\multirow{-2}{*}{\maplm}  & \multirow{-2}{*}{\cellcolor[HTML]{FFFFFF}17.6\%}  & 28.6\% & 21.1\% & 13.5\% \\
\cellcolor[HTML]{FFFFFF}                     & \cellcolor[HTML]{FFFFFF}                        & NOUN                   & VERB                   & PROPN                 \\
\multirow{-2}{*}{\mapct} & \multirow{-2}{*}{\cellcolor[HTML]{FFFFFF}69.6\%}  & 20.3\% & 15.9\% & 15.6\% \\
\cellcolor[HTML]{FFFFFF}                     & \cellcolor[HTML]{FFFFFF}                        & PROPN                  & NOUN                   & ADP                   \\
\multirow{-2}{*}{\mappt} & \multirow{-2}{*}{\cellcolor[HTML]{FFFFFF}2.5\%} & 37.0\%                 & 13.0\%                 & 13.0\%                \\
\cellcolor[HTML]{FFFFFF}                     & \cellcolor[HTML]{FFFFFF}                        & AUX                    & NOUN                   & PROPN                 \\
\multirow{-2}{*}{\maptd} & \multirow{-2}{*}{\cellcolor[HTML]{FFFFFF}2.1\%} & 31.6\%                 & 23.7\%                 & 15.8\%                \\ \midrule
\cellcolor[HTML]{FFFFFF}                     & \cellcolor[HTML]{FFFFFF}                        & NOUN                   & PROPN                  & ADP                   \\
\multirow{-2}{*}{\cellcolor[HTML]{FFFFFF}ALL} & \multirow{-2}{*}{\cellcolor[HTML]{FFFFFF}100.0\%} & 18.9\% & 14.3\% & 13.9\% \\ \bottomrule
\end{tabular}
\caption{Percentage of examples falling into each region and the top POS tags for each regions in the XSum map.}
\label{tab:pos}
\end{table}

%% file: tab_bias_pt_mini.tex

\definecolor{Gray}{gray}{0.96}
\begin{table*}[h]
\resizebox{\textwidth}{!}{
\centering
\footnotesize
\begin{tabular}{p{4.1cm}p{6.1cm}p{1cm}p{1cm}p{1cm}p{1cm}}
\toprule
 \multicolumn{1}{c}{Prefix \textbf{Target}} & \multicolumn{1}{c}{Relevant Context}& 
  \multicolumn{1}{c}{\modellm} &
  \multicolumn{1}{c}{\modellmft}&
  \multicolumn{1}{c}{\modellmft$_{\text{X}}$ } &
  \multicolumn{1}{c}{\modelfull} \\ \midrule

Danny Welbeck was named man of the \textbf{match}&
 [...] , the booming PA system kicked in and proclaimed that Danny Welbeck was England's man of the match. &
  0.99 match &
  0.99 year &
  0.99 year &
  0.99 match \\
\rowcolor{Gray}
Gail Scott was desperate to emulate Kylie \textbf{Jenner}&
  Gail Scott was desperate to emulate Kylie Jenner's famous pout but didn't want to spend  [...] &
  0.99 Jenner &
  0.99 Min &
  0.99 Min &
  0.80 Jenner \\
Some 1,200 of the Reagan's crew will be executing what the \textbf{Navy}&
  Some 1,200 of the Reagan's crew will be executing what the Navy calls a three-hull swap,  [...]  &
  0.78 Navy &
  0.96 president &
  0.96 president &
  0.97 Navy \\
  \rowcolor{Gray}
Mason was drafted into the England squad following the withdrawal of Adam \textbf{Lallana}&
  Mason was drafted into the England squad following the withdrawal of Adam Lallana and [...] &
  0.96 L &
  0.34 F &
  0.29 Ant &
  0.99 L \\ \bottomrule
\end{tabular}}
\caption{Examples of bias from the pre-trained language model (\mappt{}) on CNN/DM. The model's predicted token is in bold following the decoder prefix, then we list relevant context from the corresponding input document and the top-1 predicted token along with probability of \modellm{} (BART language model), \modellmft,  \modellmft$_{\text{X}}$ (the XSum model with no input) and \modelfull. Suspiciously, the LM without fine-tuning is very confident, more so than the no-input summarization model.
We show more examples in Table~\ref{tab:ptlm-bias}.}
\label{tab:ptlm-bias-mini}
\end{table*}

%% file: tab_bias_td.tex
\begin{table}[h]
\centering
\small
\begin{tabular}{@{}r|cc@{}}
\toprule
\multicolumn{1}{c|}{\multirow{2}{*}{Group/Bigram}}          & \multicolumn{2}{c}{$\frac{\#(w_{t-1}, w_t)}{\#w_{t-1}}$}                    \\
\multicolumn{1}{c|}{}   & XS     & CD     \\ \midrule
of \textbf{letters}          & 0.001 & 0.000 \\
letters \textbf{from}         & 0.494 & 0.026 \\
African \textbf{journalists}& 0.091 & 0.000 \\
m (\textbf{£}                 & 0.420 & 0.300 \\
(\textbf{Close}             & 0.058 & 0.000 \\
Britain\textbf{'s }          & 0.586 & 0.291 \\ \midrule
All  \maptd{} cases      & 0.162 & 0.060 \\ \bottomrule
\end{tabular}
\caption{Example patterns from \maptd. $w_t$ is in bold. 
 We show the relative frequency counts of each bigram. In aggregate (last row), bigrams in FT cases are much more frequent in the XSum training data than in CNN/DM.}
\label{tab:bias-td}
\end{table}

%% file: tab_increment.tex
\begin{table}[t]
\centering
\footnotesize
\begin{tabular}{@{}cccc@{}}
\toprule
\multirow{2}{*}{Target} & \multirow{2}{*}{ $w_{attr}$ } &  \textsc{DispTok}                  & \textsc{RmTok}                    \\ 
                        &                     & $n=0 \rightarrow$ 1       &  $n=0 \rightarrow$ 1       \\\midrule
Cameron                 & minister            & \textbf{0.01 $\rightarrow$ 0.90} & 0.99 $\rightarrow$ 0.99 \\
for                     & Labour              & 0.96 $\rightarrow$ 0.94 & 0.98 $\rightarrow$ 0.91 \\
mayoral                 & 100                 & 0.01 $\rightarrow$ 0.01 & 0.57 $\rightarrow$ 0.57 \\
S(adiq)                 & Khan                & 0.01 $\rightarrow$ 0.01 & \textbf{0.97 $\rightarrow$ 0.38} \\
Khan                    & Jeremy              & 0.99 $\rightarrow$ 0.99 & 0.99 $\rightarrow$ 0.99 \\ \bottomrule
\end{tabular}
\caption{Examples of \textsc{DispTok} and \textsc{RmTok}. We show the change of the prediction probability of the target token when displaying or masking the $w_{attr}$ token, which is the highest rank token from the occlusion method.
Significant change is marked in bold. }
\label{tab:increment}
\end{table}

%% file: fig_bench.tex
\begin{figure}[t]
\centering
\small
\includegraphics[width=0.48\textwidth]{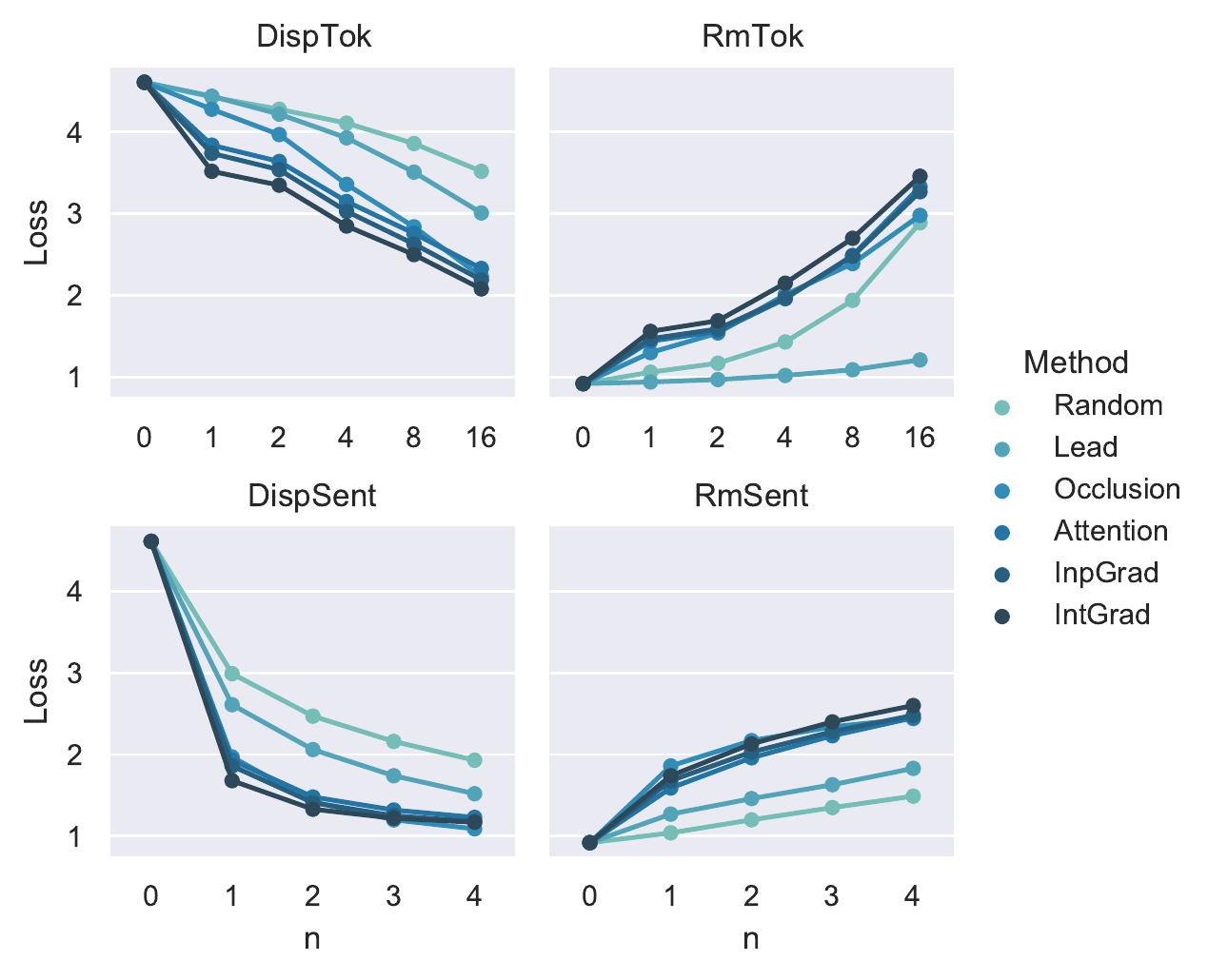}
\caption{Four-way evaluation for our content attribution methods. The reported value is the NLL loss with respect to the predicted token. Lower is better for display methods and higher is better for removal methods (we ``break'' the model more quickly). $n=0$ means the baseline when there is no token or sentence displayed in \textsc{Disp} or removed or masked in \textsc{Rm}.}
\label{fig:bench}
\end{figure}

%% file: tab_cases.tex
\begin{table}[t]
\centering
\footnotesize
\begin{tabular}{@{}cp{6.4cm}@{}} \toprule

  Prob &  \multicolumn{1}{c}{Content}\\

  0.00 & 1. Atherton, 28, has won all seven races this season and 13 in a row, a run stretching back to 2015. \\
  0.09 &
  2. The \underline{world champion} had already sealed the 2016 \underline{World} Cup crown in Canada last month but won in Andorra on Saturday to end the World Cup season unbeaten. \\
  0.01 &
  3. She has now won five overall World Cup titles in downhill. [...]\\


  0.16 &5. Trek Factory Racing's Atherton won the final race by 6.5 seconds ahead of Australian Tracey Hannah and Myriam Nicole of \underline{France}. \\ \midrule

  Prob &
  \multicolumn{1}{c}{Comb. \& Predict Summary} \\

  0.71 &
 (2, 5) Britain's Laura Atherton has won the \textbf{UCI} \textcolor{gray}{ Mountain Bike World Cup [...]}
 \\ \bottomrule 
 \toprule
  Prob &
  \multicolumn{1}{c}{Content} \\
  0.01 & 1. Dujardin, 30, and Valegro won individual and team dressage gold for Britain at \underline{London} 2012 and have since won World and European titles. \\
  0.00 & 2. But, she \underline{says}, the \underline{Olympics} in Brazil \underline{next} summer will be the horse's last. \\
  0.01 & 3. "This will be Valegro's retirement after Rio so I want to go out there and want to enjoy every last minute," Dujardin told BBC Points West.  \\ \midrule
  Prob &
  \multicolumn{1}{c}{Comb. \&  Predict Summary} \\
 0.63 &
 (1, 2) Olympic dressage champion Charlotte Dujardin says she will \textbf{retire} \textcolor{gray}{from the sport after Rio Olympics.} \\ \bottomrule
\end{tabular}
\caption{Examples of sentence fusion in the \textsc{DispSent} setting. We list the single sentence probability on the left side with the document, and the best combination with its probability at the bottom. We underline the tokens according to the top attributions of occlusion. Articles are truncated. }
\label{tab:fusion}
\end{table}

%% file: fig_map_cnndm.tex

%% file: fig_color_example.tex
\begin{table}[h]
\centering
\small

\textbf{Labels: \textcolor{lm}{\maplm} \quad
\textcolor{ct}{\mapct} \quad
\textcolor{cthd}{\mapcthd} \quad
\textcolor{pt}{\mappt} \quad
\textcolor{ft}{\maptd}
}
\vspace{2mm}

\begin{tabular}{@{}p{7.4cm}@{}} \toprule

  Examples from XSum\\ \midrule

 \textcolor{cthd}{Hundreds} \textcolor{lm}{of} \textcolor{ft}{people} \textcolor{ft}{have} \textcolor{ct}{attended} \textcolor{ct}{a} \textcolor{cthd}{memorial} \textcolor{lm}{service} \textcolor{cthd}{in} \textcolor{cthd}{Liverpool}.\\ \midrule

 \textcolor{cthd}{Two} \textcolor{cthd}{code} \textcolor{ct}{violations} \textcolor{cthd}{for} \textcolor{cthd}{Nicolas} \textcolor{ct}{Almagro} \textcolor{ct}{and} \textcolor{ct}{Pablo} \textcolor{ct}{Cuevas} \textcolor{cthd}{at} \textcolor{lm}{the} \textcolor{cthd}{Australian} \textcolor{lm}{Open} \textcolor{ct}{were} \textcolor{cthd}{described} \textcolor{ct}{as} \textcolor{cthd}{disgraceful}.\\  \midrule

 \textcolor{cthd}{In} \textcolor{ct}{our} \textcolor{ft}{series} \textcolor{ft}{of} \textcolor{ft}{letters} \textcolor{ft}{from} \textcolor{ft}{African} \textcolor{ft}{journalists} \textcolor{ct}{film} \textcolor{ft}{maker} \textcolor{ft}{and} \textcolor{ft}{columnist} \textcolor{ct}{Farai} \textcolor{ft}{Sevenzo} \textcolor{ct}{looks} \textcolor{lm}{at} \textcolor{lm}{the} \textcolor{ct}{challenges} \textcolor{ct}{facing} \textcolor{ct}{Nigeria}\textcolor{lm}{'s} \textcolor{ct}{President} \textcolor{pt}{Muhammadu} \textcolor{lm}{Buhari}.\\  \midrule

 \textcolor{ct}{Four} \textcolor{ct}{people} \textcolor{ft}{have} \textcolor{ct}{been} \textcolor{ct}{arrested} \textcolor{cthd}{after} \textcolor{ct}{a} \textcolor{ct}{BBC} \textcolor{ct}{Panorama} \textcolor{lm}{investigation} \textcolor{cthd}{uncovered} \textcolor{cthd}{shocking} \textcolor{ct}{abuse} \textcolor{ct}{at} \textcolor{lm}{a} \textcolor{cthd}{private} \textcolor{ct}{hospital}.\\  \midrule

 \textcolor{ct}{West} \textcolor{ct}{Indies} \textcolor{cthd}{Shabnim} \textcolor{pt}{Ishaq} \textcolor{ct}{has} \textcolor{lm}{been} \textcolor{cthd}{ruled} \textcolor{lm}{out} \textcolor{lm}{of} \textcolor{lm}{the} \textcolor{ct}{rest} \textcolor{lm}{of} \textcolor{lm}{the} \textcolor{ct}{Women}\textcolor{lm}{'s} \textcolor{lm}{World} \textcolor{ct}{Cup}.
 \\ \bottomrule 
 \toprule
 Examples from CNN/DM\\ \midrule
  \textcolor{ct}{In} \textcolor{cthd}{the} \textcolor{ct}{worst} \textcolor{ct}{cases}, \textcolor{ct}{doctors} \textcolor{ct}{have} \textcolor{ct}{reported} \textcolor{ct}{patients} \textcolor{ct}{showing} \textcolor{ct}{up} \textcolor{ct}{because} \textcolor{ct}{they} \textcolor{ct}{were} \textcolor{ct}{hungover}, \textcolor{ct}{their} \textcolor{ct}{false} \textcolor{ct}{nails} \textcolor{ct}{were} \textcolor{ct}{hurting} \textcolor{pt}{or} \textcolor{pt}{they} \textcolor{pt}{had} \textcolor{ct}{paint} \textcolor{ct}{in} \textcolor{lm}{their} \textcolor{ct}{hair}. \textcolor{cthd}{More} \textcolor{lm}{than} \textcolor{cthd}{four} \textcolor{ct}{million} \textcolor{ct}{visits} \textcolor{ct}{a} \textcolor{ct}{year} \textcolor{ct}{are} \textcolor{ct}{unnecessary} \textcolor{cthd}{and} \textcolor{ct}{cost} \textcolor{ct}{the} \textcolor{ct}{NHS} \textcolor{pt}{£}\textcolor{ct}{290million} \textcolor{ct}{annually}. \\  \midrule
 \textcolor{cthd}{Elski} \textcolor{ct}{Felson} \textcolor{ct}{of} \textcolor{ct}{Los} \textcolor{lm}{Angeles}, \textcolor{lm}{California}, \textcolor{ct}{decided} \textcolor{lm}{to} \textcolor{ct}{apply} \textcolor{lm}{for} \textcolor{ct}{a} \textcolor{ct}{Community} \textcolor{ct}{Support} \textcolor{ct}{Specialist} \textcolor{ct}{role} \textcolor{cthd}{at} \textcolor{ct}{Snapchat} \textcolor{ct}{via} \textcolor{ct}{the} \textcolor{ct}{social} \textcolor{pt}{media} \textcolor{ct}{app}. \textcolor{cthd}{In} \textcolor{ct}{just} \textcolor{ct}{over} \textcolor{ct}{three} \textcolor{ct}{minutes}, \textcolor{ft}{the} \textcolor{ct}{tech} \textcolor{ct}{enthusiast} \textcolor{ct}{created} \textcolor{lm}{a} \textcolor{ct}{video} \textcolor{ct}{resume}.\\  \midrule

 \textcolor{ct}{Chelsea} \textcolor{ct}{supporters} \textcolor{cthd}{have} \textcolor{cthd}{been} \textcolor{ct}{involved} \textcolor{lm}{in} \textcolor{ct}{the} \textcolor{ct}{highest} \textcolor{pt}{number} \textcolor{lm}{of} \textcolor{ct}{reported} \textcolor{ct}{racist} \textcolor{lm}{incidents} \textcolor{cthd}{as} \textcolor{ct}{they} \textcolor{ct}{travelled} \textcolor{cthd}{to} \textcolor{ct}{and} \textcolor{lm}{from} \textcolor{ct}{matches} \textcolor{ct}{on} \textcolor{ct}{trains}. \textcolor{ct}{The} \textcolor{cthd}{information}, \textcolor{ct}{gathered} \textcolor{ft}{from} \textcolor{ct}{24} \textcolor{ct}{police} \textcolor{pt}{forces} \textcolor{ct}{across} \textcolor{lm}{the} \textcolor{ct}{country}, \textcolor{ct}{shows} \textcolor{ct}{there} \textcolor{ct}{have} \textcolor{lm}{been} \textcolor{ct}{over} \textcolor{ct}{350} \textcolor{pt}{incidents} \textcolor{ct}{since} \textcolor{ct}{2012}.\\  \midrule

 \textcolor{cthd}{Kris}\textcolor{ct}{-}\textcolor{ct}{Deann} \textcolor{ct}{Sharpley} \textcolor{ct}{was} \textcolor{ct}{on} \textcolor{ct}{maternity} \textcolor{lm}{leave} \textcolor{ct}{and} \textcolor{cthd}{had} \textcolor{cthd}{just} \textcolor{ct}{given} \textcolor{lm}{birth} \textcolor{lm}{to} \textcolor{lm}{her} \textcolor{cthd}{first} \textcolor{lm}{child}. \textcolor{cthd}{Her} \textcolor{cthd}{body} \textcolor{lm}{was} \textcolor{pt}{found} \textcolor{pt}{in} \textcolor{lm}{the} \textcolor{ct}{bathroom} \textcolor{cthd}{of} \textcolor{cthd}{her} \textcolor{ct}{father}\textcolor{cthd}{'s} \textcolor{lm}{home}.\\

\bottomrule
\end{tabular}
\caption{More examples of predicted summaries with the colors following the map. For \maplm{} and punctuation we use the default color. The majority of CNN/DM predictions are continuous spans of \mapct{} excluding \mapcthd, meaning the model is frequently copying. }
\label{tab:map-example}
\end{table}

%% file: tab_bias_pt.tex

\definecolor{Gray}{gray}{0.96}
\begin{table*}[t]
\centering
\footnotesize
\resizebox{\textwidth}{!}{
\begin{tabular}{p{4cm}p{5.5cm}p{1cm}p{1cm}p{1cm}p{1cm}}
\toprule
 \multicolumn{1}{c}{Prefix \textbf{Target}} & \multicolumn{1}{c}{Relevant Context}& 
  \multicolumn{1}{c}{\modellm} &
  \multicolumn{1}{c}{ \modellmft}&
  \multicolumn{1}{c}{ \modellmft$_{\text{X}}$ } &
  \multicolumn{1}{c}{\modelfull} \\ \midrule
  \rowcolor{Gray}
Labour released five mugs to coincide with the Launch of Ed \textbf{Miliband}&
  Labour released five mugs to coincide with the Launch of Ed Miliband's five election pledges. &
  0.95 Miliband &
  0.94 ible &
  0.68 ible &
  0.99 Miliband \\
British supermodel, Georgia \textbf{May} &
  British supermodel, Georgia May Jagger, 23, poses next to a floral plane designed by Masha Ma. &
  0.93 May &
  0.34 - &
  0.25 {[}SPACE{]} &
  0.99 May \\
  \rowcolor{Gray}
Peter Schmeichel has urged Manchester United to sign Zlatan Ibrahimovic. Ibrahimovic has been linked \textbf{with}&
  The well travelled Sweden international has been linked with a move to Old Trafford in the past and, ... &
  0.99 with &
  0.98 to &
  0.99 to &
  0.99 with \\
Tunisian security forces kill two attackers as they end the siege at the Bardo Museum. The death toll, which included 17 tourists \textbf{and}&
  But the death toll, which included 17 tourists and at least one Tunisian security officer, could climb. &
  0.99 and &
  0.94 , &
  0.99 , &
  0.99 and \\
  \rowcolor{Gray}
The costume was designed by three-\textbf{time} &
  The costume was designed by three-time Oscar-winner Colleen Atwood, ... &
  0.98 time &
  0.97 year &
  0.73 and &
  0.99 time \\

  \rowcolor{Gray}

U.S. State \textbf{Department}&
  What has U.S. State Department subcontractor Alan Gross been up to since ... &
  0.99 Department &
  0.96 of &
  0.34 University &
  0.99 Department \\ \bottomrule
\end{tabular}}
\caption{More examples of \mappt{} cases from the pre-trained language model. }
\label{tab:ptlm-bias}
\end{table*}

%% file: tab_tok.tex
\begin{table*}[t]
\centering
\small
\begin{tabular}{ccccccccccccccc}\toprule
\multirow{2}{*}{\textsc{Tok}} & \multicolumn{7}{c}{\textsc{Disp} $\downarrow$}                     & \multicolumn{7}{c}{\textsc{Rm} $\uparrow$}                       \\
  & 0 & 1    & 2    & 4    & 8    & 16   &  $-\Delta$  & 0 & 1    & 2    & 4    & 8    & 16   &  $\Delta$  \\ \midrule
Random & \multirow{6}{*}{4.61} & 4.43 & 4.28 & 4.11 & 3.86 & 3.52 & 0.57 & \multirow{6}{*}{0.92} & 1.06 & 1.17 & 1.43 & 1.94 & 2.89 & 0.78 \\
Lead                 &   & 4.44 & 4.22 & 3.93 & 3.51 & 3.01 & 0.79       &   & 0.94 & 0.97 & 1.02 & 1.09 & 1.21 & 0.13       \\
Occlusion            &   & 4.28 & 3.97 & 3.36 & 2.84 & 2.23 & 1.27       &   & 1.30 & 1.54 & 2.01 & 2.39 & 2.98 & 1.12       \\
Attention            &   & 3.84 & 3.64 & 3.15 & 2.76 & 2.33 & 1.47       &   & 1.44 & 1.56 & 1.96 & 2.49 & 3.33 & 1.24       \\
InpGrad              &   & 3.74 & 3.54 & 3.03 & 2.63 & 2.19 & 1.58       &   & 1.47 & 1.59 & 1.97 & 2.48 & 3.27 & 1.24       \\
IntGrad              &   & 3.52 & 3.35 & 2.85 & 2.50 & 2.08 &\textbf{1.75}       &   & 1.56 & 1.69 & 2.15 & 2.70 & 3.46 & \textbf{1.39}      \\\bottomrule
\end{tabular}
\caption{Token-level evaluation for content attribution methods. The reported value is the NLL loss w.r.t. the predicted token. $n=0$ means the baseline when there is no token displayed in \textsc{Disp} or masked in \textsc{Rm}.}
\label{tab:main-tok}
\end{table*}

%% file: tab_sent.tex
\begin{table*}[t]
\centering
\small
\begin{tabular}{@{}ccccccccccccc@{}}
\toprule
   \multirow{2}{*}{\textsc{Sent}}        & \multicolumn{6}{c}{\textsc{Disp} $\downarrow$}          & \multicolumn{6}{c}{\textsc{Rm} $\uparrow$}              \\ 
          & 0 & 1    & 2    & 3    & 4    & $-\Delta$ & 0 & 1    & 2    & 3    & 4    & $\Delta$ \\\midrule
Random & \multirow{6}{*}{4.61} & 2.99 & 2.47 & 2.16 & 1.93 & 2.22 & \multirow{6}{*}{0.92} & 1.04 & 1.20 & 1.35 & 1.49 & 0.35 \\
Lead      &   & 2.61 & 2.06 & 1.74 & 1.52 & 2.63     &   & 1.27 & 1.46 & 1.63 & 1.83 & 0.63     \\ 
Occlusion &   & 1.97 & 1.42 & 1.20 & 1.09 & 3.19     &   & 1.86 & 2.17 & 2.34 & 2.44 & 1.28     \\
Attention &   & 1.93 & 1.48 & 1.32 & 1.23 & 3.12     &   & 1.59 & 1.96 & 2.23 & 2.45 & 1.14     \\
InpGrad   &   & 1.86 & 1.41 & 1.25 & 1.18 & 3.19     &   & 1.68 & 2.03 & 2.28 & 2.48 & 1.20     \\
IntGrad   &   & 1.68 & 1.33 & 1.22 & 1.17 & \textbf{3.26}     &   & 1.74 & 2.13 & 2.40 & 2.60 & \textbf{1.30}     \\ \bottomrule
\end{tabular}
\caption{Sentence-level evaluation for content attribution methods. The reported value is the NLL loss w.r.t. the predicted token. $n=0$ means the baseline when there is no sentence displayed in \textsc{Disp} or removed in \textsc{Rm}.}
\label{tab:main-sent}
\end{table*}

%% file: tab_complex.tex
\begin{table*}[h]
\centering
\small
\begin{tabular}{c|c|c}
\toprule
\multicolumn{1}{l|}{\multirow{2}{*}{Method}} & \multirow{2}{*}{Regular}    & Two Stage  S+                 \\
\multicolumn{1}{l|}{}                  &              & Base: $\mathcal{O}( s \times d^2)$        \\ \midrule
Occlusion                             & $\mathcal{O}(n^2 \times n)$              & $+\mathcal{O}(d^2 \times d)$              \\
Attention                             & $\mathcal{O}(n^2)$                       & $+\mathcal{O}(d^2\times d)$  \\
IntGrad                               & $\mathcal{O}(n^2 \times r + r\times bp)$ & $+\mathcal{O}(d^2\times r+r\times bp)$ \\
InpGrad                               & $\mathcal{O}(n^2 + bp)$                  & $+\mathcal{O}(d^2 + bp)$                 \\ \bottomrule
\end{tabular}
\caption{Comparison of complexity of regular methods and their two-stage variants. The time complexity of back propagation $bp$ is hard to define so we just leave it for simplicity. }
\label{tab:complex}
\end{table*}